\documentclass[letterpaper]{article} % DO NOT CHANGE THIS
\usepackage{aaai2026}  % DO NOT CHANGE THIS
\usepackage{times}  % DO NOT CHANGE THIS
\usepackage{helvet}  % DO NOT CHANGE THIS
\usepackage{courier}  % DO NOT CHANGE THIS
\usepackage[hyphens]{url}  % DO NOT CHANGE THIS
\usepackage{graphicx} % DO NOT CHANGE THIS
\usepackage{natbib}  % DO NOT CHANGE THIS AND DO NOT ADD ANY OPTIONS TO IT
\usepackage{caption} % DO NOT CHANGE THIS AND DO NOT ADD ANY OPTIONS TO IT
\usepackage{algorithm}
\usepackage{algorithmic}

\usepackage{amsmath}
\usepackage{amsfonts}
\usepackage{amssymb}

\usepackage{xcolor}
\usepackage{color}

\usepackage{subcaption} % Add this in your preamble

\usepackage{textpos} % for absolute positioning
\usepackage{booktabs}
\usepackage{verbatim}
\usepackage{graphicx}
\usepackage[T1]{fontenc}
\usepackage{graphicx}
\usepackage{newfloat}
\usepackage{listings}
\DeclareCaptionStyle{ruled}{labelfont=normalfont,labelsep=colon,strut=off} % DO NOT CHANGE THIS
\floatstyle{ruled}
\newfloat{listing}{tb}{lst}{}
\floatname{listing}{Listing}
\title{AF-MAT: Aspect-aware Flip-and-Fuse xLSTM for Aspect-based Sentiment Analysis}

\author {
    Adamu Lawan\textsuperscript{\rm 1},
    Juhua Pu\textsuperscript{\rm 1},
    Haruna Yunusa\textsuperscript{\rm 2},
    Muhammad Lawan\textsuperscript{\rm 3},
    Mahmoud Basi\textsuperscript{\rm 4},
    \textbf{Muhammad Adam}\textsuperscript{\rm 5}
}
\affiliations {
    \textsuperscript{\rm 1}School of Computer Science and Technology, Beihang University, Beijing, China\\
    \textsuperscript{\rm 2}School of Automation Science and Electrical Engineering, Beihang University, Beijing, China\\
    \textsuperscript{\rm 3}Department of Information and Communication Technology, Federal University, Gusau, Nigeria\\
    \textsuperscript{\rm 4}School of Cyber Science and Technology, Beihang University, Beijing, China\\
    \textsuperscript{\rm 5}Department of Computer \& Information Sciences, Maryland, United States\\
    \texttt{\{alawan, pujh, yunusa2k2\}@buaa.edu.cn}
}

\usepackage{bibentry}
\begin{document}

\maketitle

\begin{abstract}
Aspect-based Sentiment Analysis (ABSA) is a crucial NLP task that extracts fine-grained opinions and sentiments from text, such as product reviews and customer feedback. Existing methods often trade off efficiency for performance: traditional LSTM or RNN models struggle to capture long-range dependencies, transformer-based methods are computationally costly, and Mamba-based approaches rely on CUDA and weaken local dependency modeling. The recently proposed Extended Long Short-Term Memory (xLSTM) model offers a promising alternative by effectively capturing long-range dependencies through exponential gating and enhanced memory variants, sLSTM for modeling local dependencies, and mLSTM for scalable, parallelizable memory. However, xLSTM’s application in ABSA remains unexplored. To address this, we introduce Aspect-aware Flip-and-Fuse xLSTM (AF-MAT), a framework that leverages xLSTM’s strengths. AF-MAT features an Aspect-aware matrix LSTM (AA-mLSTM) mechanism that introduces a dedicated aspect gate, enabling the model to selectively emphasize tokens semantically relevant to the target aspect during memory updates. To model multi-scale context, we incorporate a FlipMix block that sequentially applies a partially flipped Conv1D (pf-Conv1D) to capture short-range dependencies in reverse order, followed by a fully flipped mLSTM (ff-mLSTM) to model long-range dependencies via full sequence reversal. Additionally, we propose MC2F, a lightweight Multihead Cross-Feature Fusion based on mLSTM gating, which dynamically fuses AA-mLSTM outputs (queries and keys) with FlipMix outputs (values) for adaptive representation integration. Experiments on three benchmark datasets demonstrate that AF-MAT outperforms state-of-the-art baselines, achieving higher accuracy in ABSA tasks.
\end{abstract}


\section{Introduction}

\begin{figure}[h!]
    \centering
    \fbox{
        \includegraphics[width=0.8\linewidth]{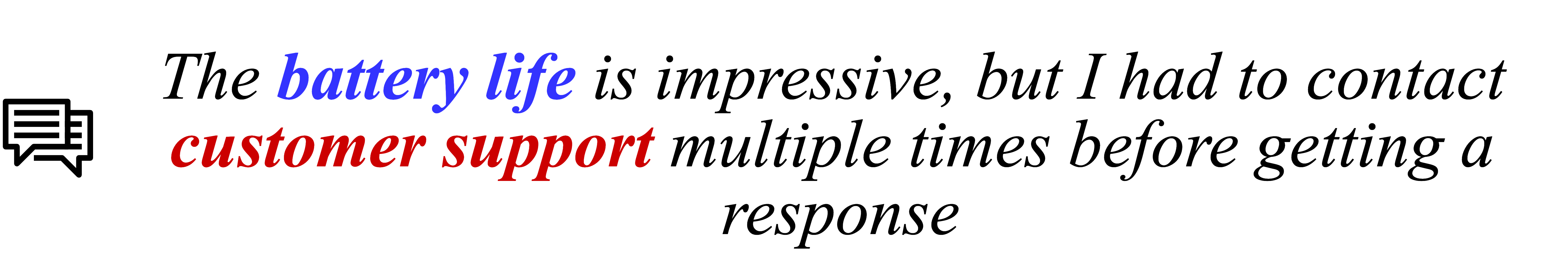}
    }
    \caption{An example sentence demonstrating the need for both short- and long-range aspect-sentiment dependencies. The positive sentiment for \textit{battery life} is derived from immediate contextual cues, while the negative sentiment for \textit{customer support} emerges from a broader context involving delayed response. This highlights a key challenge in ABSA: accurately associating sentiments with their corresponding aspects when contrasting opinions span varying contextual distances within a single sentence.}
    \label{fig:short_long_dependency}
\end{figure}

Aspect-based Sentiment Analysis (ABSA) is an essential Natural Language Processing (NLP) task that identifies sentiments for specific aspects, such as product features, enabling targeted sentiment analysis. Figure \ref{fig:short_long_dependency} illustrate ABSA in details.
% -----------------------
% "Image-like" colored sentence
Previous ABSA models rely on attention mechanisms \cite{Tang2016, Wang2016, Ma2017, Peng2017, Tay2018, hazarika2018, Fan2018, Song2019, Yang2019, Liu2020, Yadav2021, Wang2021} to capture aspect-sentiment associations, achieving strong context modeling performance. However, these models suffer from quadratic complexity with increasing sequence length and are sensitive to noise in complex or informal sentences such as in social media text, limiting their efficiency and robustness. To address attention-based limitations subsequent work in ABSA has shifted toward syntactic approaches. These methods ~\cite{Sun2019, Zhang2019b, li-etal-2021-dual-graph, Liang2022, zhang-etal-2022-ssegcn, Gu2023a, Wu2023, Liu2023, Li2023, Ouyang2024, Wu2026, Feng2022, Yu2025} utilize Graph Convolutional Networks (GCNs) to leverage dependency-based structural information and external knowledge, effectively capturing syntactic relationships between aspects and context words for improved aspect extraction and sentiment classification. Recent advancements in ABSA leverage GCNs and Selective State Space model (Mamba) \cite{Gu2023} to effectively capture long-range dependencies between aspect and opinion words, enhancing sentiment analysis precision \cite{lawan-etal-2025-enhancing}.

Despite progress in ABSA, effectively capturing long-range dependencies while preserving localized aspect-specific cues remains a core challenge. Attention mechanisms suffer from quadratic complexity, and Mamba-based models face CUDA dependency, limiting scalability and efficiency. The recently introduced Extended Long Short-Term Memory (xLSTM) model~\cite{Beck2024} has shown promising progress in efficiently capturing long-range dependencies in NLP, due to its innovative architecture, which combines exponential gating with enhanced memory variants, scalar LSTM (sLSTM) and matrix LSTM (mLSTM). The gating mechanism enables precise control over long-range information flow, while mLSTM introduces a parallelizable matrix memory for better efficiency and scalability. However, its potential for ABSA, particularly in capturing critical aspect-sentiment associations, remains untapped. Unlike prior attention- or GCN-based methods that rely on quadratic-time mechanisms or error-prone syntactic dependencies, our approach leverages the efficient mLSTM architecture to model multi-scale aspect-sentiment interactions within a modular, linear-time framework.

To this end, we propose Aspect-aware Flip-and-Fuse xLSTM (AFMAT), a novel framework based on xLSTM that introduces several specialized modules for precise and efficient ABSA. At its core, AF-MAT introduces an Aspect-aware mLSTM (AA-mLSTM), which integrates a dedicated aspect gate to selectively emphasize tokens that are semantically aligned with the target aspect during memory updates. To improve multi-scale context modeling, we introduce a novel FlipMix module that sequentially applies two transformations: a partially flipped Conv1D (pf-Conv1D) module that preserves short-range aspect-sentiment dependencies in reverse order, and a fully flipped mLSTM (ff-mLSTM) module that captures long-range context through complete sequence reversal. Furthermore, we design MC2F, a lightweight Multihead Cross-Feature Fusion mechanism that employs mLSTM gating to dynamically integrate the outputs of AA-mLSTM (as queries and keys) with FlipMix outputs (as values). Unlike conventional attention-based fusion, MC2F enables efficient, adaptive, and linear-time cross-feature interaction, enhancing both computational efficiency and representational power.
The main contributions of this paper are summarized as follows:
\begin{itemize}
\item We present the first integration of the xLSTM architecture into the ABSA domain, introducing the novel AF-MAT framework, which effectively models aspect-oriented dependencies and enables rich contextual interaction.
\item We propose three novel modules, AA-mLSTM, FlipMix, and MC2F designed to capture short and long-range aspect-sentiment relationships from reverse-oriented sequences, and to fuse multi-scale contextual representations efficiently.
\item We validate AF-MAT’s performance on three public benchmark datasets, demonstrating its superiority.
\end{itemize}

\section{Related Work}

ABSA has progressed significantly with advanced architectures like transformer-based models \cite{Tang2020} and GCNs \cite{Sun2019}. Transformers excel in ABSA by capturing long-range dependencies via self-attention, enabling precise sentiment analysis. For instance, MemNet \cite{Tang2016} employs a deep memory network with multiple attention layers over external memory to model context word importance effectively. Other attention-based models, including IAN \cite{Ma2017}, RAM \cite{Peng2017}, and MGAN \cite{Fan2018}, enhance ABSA by leveraging interactive or multi-granularity attention for targeted sentiment tasks.

In contrast, GCN-based models for ABSA enhance aspect-context relationships by leveraging GCNs that operate on a sentence’s dependency tree to model syntactic and semantic interactions. For example, EK-GCN \cite{Gu2023a} integrates sentiment lexicons, part-of-speech matrices, and a word–sentence Interaction Network to strengthen context-aspect interactions, improving sentiment classification accuracy. Other notable GCN-based models, including ASGCN \cite{Zhang2019b}, AG-VSR \cite{Feng2022}, KHGCN \cite{Song2024}, and ASHGAT \cite{Ouyang2024}, further advance ABSA by incorporating adaptive graph structures, or knowledge-enhanced GCNs, achieving robust performance. State space models (SSMs) \cite{Gu2020, Gu2021, Gu2023} have risen to prominence in NLP for their robust handling of long-range dependencies, enabling effective modeling of complex sequence interactions. Lawan et al. \cite{lawan-etal-2025-enhancing} integrate GCNs with Mamba’s SSM framework to capture complex dependencies between aspect and opinion words, significantly improving sentiment analysis precision on the ABSA benchmark datasets.

Relatedly, xLSTM \cite{Beck2024} applications in time series forecasting \cite{Kong2025}, remote sensing \cite{Wu2024}, wind forecasting \cite{He2025}, computer vision \cite{alkin2025}, and speech enhancement \cite{Khne2025}, and bidirectional Mamba in recommendation systems \cite{Liu_2025}, speech processing \cite{10985910}, and computer vision \cite{Zhu2024}, highlight their potential for sequential tasks.

\begin{figure*}[t] % Use t for top placement
	\centering
	\includegraphics[width=\textwidth]{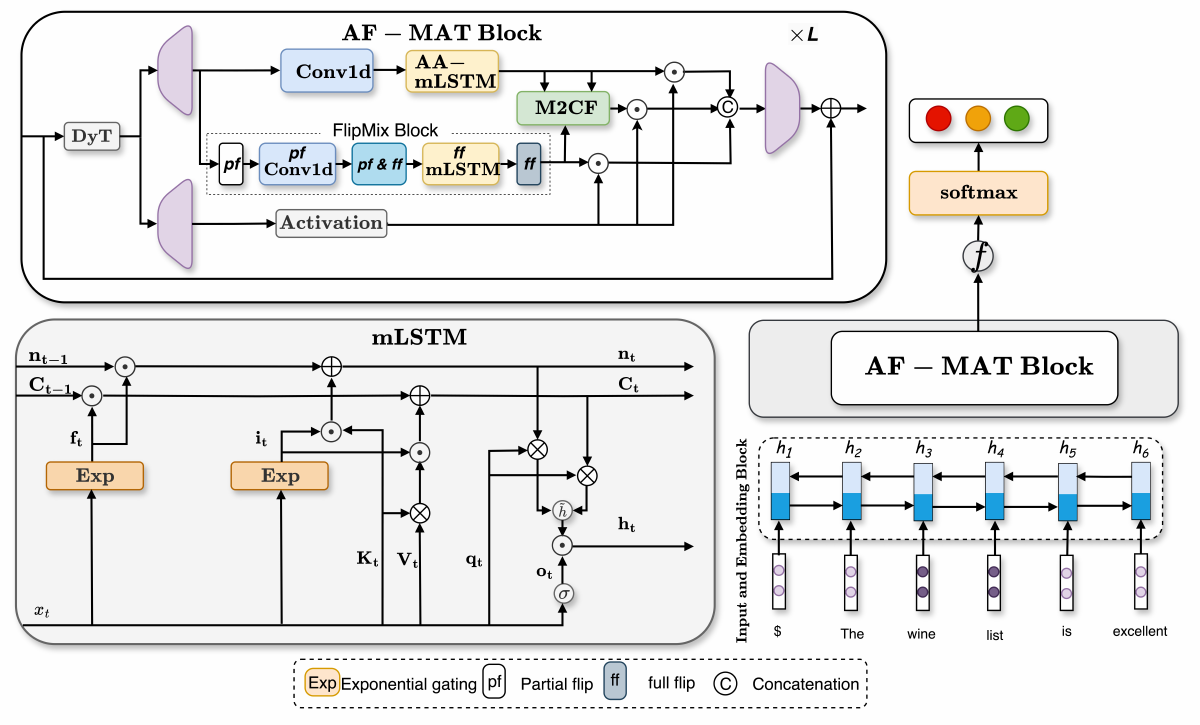}
	\caption{The complete AF-MAT architecture for ABSA. The model integrates AA-mLSTM with a dedicated aspect gate, a FlipMix module that captures multi-scale reversed dependencies via pf-Conv1D and ff-mLSTM, and an efficient MC2F mechanism. Together, these modules enable AF-MAT to model aspect-oriented, short and long-range aspect-sentiment dependencies efficiently in linear time using the mLSTM.}
	\label{fig:xlstm_fusion}
\end{figure*}

\section{Proposed AF-MAT architecture}
An overview of AF-MAT is shown in Figure \ref{fig:xlstm_fusion}. In this section, we describe the AF-MAT architecture, which is mainly composed of four components: the input and embedding block, the AA-mLSTM block, the FlipMix block, and the MC2F block. Next, components of AF-MAT architecture will be introduced separately in the rest of the sections.
\subsection{Input and Embedding Block}
For a sentence \( s = \{w_1, w_2, \ldots, w_N\} \) with an aspect \( a = \{a_1, a_2, \ldots, a_M\}  \) as a subsequence, we employ BiLSTM or BERT to encode contextual relationships. Words in \( s \) are mapped to low-dimensional vectors using an embedding matrix \( E \in \mathbb{R}^{|\mathcal{V}| \times d_e}\), where \( |\mathcal{V}| \) denotes the vocabulary size, and \( d_e \) represents the dimensionality of word embeddings, producing embeddings \( x = \{x_1, x_2, \ldots, x_N\} \). In the BiLSTM approach, these embeddings are processed to generate hidden state vectors \( H = \{h_1, h_2, \ldots, h_N\} \), capturing bidirectional context. The aspect-specific subsequence \( h_a \) is extracted from the hidden state matrix \( h_a = \{h_{a_1}, h_{a_2}, \ldots, h_{a_M}\} \). Alternatively, BERT processes the input formatted as ``[CLS] sentence [SEP] aspect [SEP],'' leveraging self-attention to model complex dependencies between aspect and opinion words, yielding contextual embeddings.
\subsection{Aspect-aware mLSTM (AA-mLSTM)}
Given an embedding sequence $H$, we begin by enhancing token representations using a Dynamic Tanh (DyT) layer, which adaptively modulates features, followed by a Linear layer to project the transformed features into a suitable space. A Conv1D layer then extracts local contextual patterns, producing a refined sequence: 
\begin{equation}
    H^{norm} = \text{SiLu}(\text{Linear}(\text{DyT}(H)))
    \label{eq:hnorm}
\end{equation}
\begin{equation}
    \tilde{H}^{fwd} = (\text{Conv1d}(\text{Linear}(\text{DyT}(H))))
    \label{eq:mlstm_out}
\end{equation}

where Conv1d(·), Linear(·) and DyT(·) is the 1-D convolution, the linear projection and the dynamic tanh \cite{Zhu_2025_CVPR} respectively.

Unlike sentence-level sentiment classification, ABSA aims to determine the sentiment toward a specific aspect term within its context sentence, requiring the modeling of semantic correlations tailored to different aspect terms. We propose an aspect-aware mLSTM that introduces a dedicated aspect gate, enabling the model to selectively emphasize tokens semantically relevant to the target aspect during memory updates. Given a contextualized input sequence $\tilde{H}^{fwd}$, we first extract the representations of aspect tokens $a$, yielding $\{\tilde{h}_{a_1}^{fwd}, \ldots, \tilde{h}_{a_M}^{fwd} \}$. We then apply mean pooling over these aspect token embeddings:

\begin{equation}
\tilde{H}_a^{fwd} = \frac{1}{M} \sum_{m=1}^M \tilde{h}_{a_m}^{fwd}
\end{equation}

The resulting vector \(\tilde{H}_a^{fwd}\) serves as the global aspect representation and is broadcast across all time steps, which is used in two ways: (1) as input to a new aspect gate, and (2) as the \emph{key} vector $k_t$ in attention-based combination :

\begin{equation}
a_t = \exp(W_a [\tilde{H_a}^{fwd} \oplus \tilde{H_a}^{fwd} \oplus \tilde{H_a}^{fwd}] + b_a) 
\end{equation}
where \(a_t\) is the aspect gate and \( \oplus \) denotes concatenation.

This gate modulates the memory update for each token by dynamically scaling the contribution of its aspect-conditioned signal. The gate is integrated into the stabilized decay matrix:

\begin{equation}
\log \mathcal{D}_{t,j} = \log \mathcal{F}_{t,j} + \log i_j + \log a_j
\end{equation}

where $\mathcal{F}_{t,j}$ denotes the cumulative forget gate from timestep $j$ to $t$, and $i_j$, $a_j$ are input and aspect gates at position $j$. By incorporating $a_j$, the model amplifies updates for positions semantically aligned with the aspect, allowing better discrimination of sentiment cues.

This enhancement enables the mLSTM block to focus its memory and attention on aspect-relevant substructures in the sequence, significantly improving the alignment between sentiment expressions and their corresponding aspects. The complete aspect-aware mLSTM forward pass is: 
\begin{equation}
\log F_{t,j} = \sum_{l=j}^{t-1} \log f_l
\end{equation}

\begin{equation}
D_{t,j} = \exp\left( \log D_{t,j} - \max_j \log D_{t,j} \right)
\end{equation}

\begin{equation}
\alpha_{t,j} = \frac{q_t^\top k_j}{\sqrt{d}}
\end{equation}

\begin{equation}
C_{t,j} = \alpha_{t,j} \cdot D_{t,j}
\end{equation}

\begin{equation}
\widetilde{C}_{t,j} = \frac{C_{t,j}}{\sum_{j=1}^t C_{t,j} + \epsilon}
\end{equation}

\begin{equation}
H_t^{fwd} = (\sum_{j=1}^{t} \widetilde{C}_{t,j} \cdot v_j) \odot H_t^{norm}
\end{equation}
Where $\epsilon$ is a small constant added for numerical stability and ($\cdot$) is scalar-vector multiplication.
We compute the \(q_t\), \(v_t\), \(i_t\), and \(f_t\) as described in Equation~\ref{eq:mlstm_fused}, using \(\tilde{H}^{\text{fwd}}\) as input.

\subsection{FlipMix Block}
Existing bidirectional xLSTM methods~\cite{Khne2025} reverse the entire sequence to model long-range dependencies, but risk weakening short-range aspect-opinion associations that are critical for ABSA. Inspired by the success of partial flipping in recommendation systems for preserving local order~\cite{Liu_2025}, we introduce a two-stage FlipMix block that first captures local patterns in reverse order, then encodes long-range semantics in a causally coherent manner. We define short-range dependencies as sentiment clues that occur in close proximity to the aspect, while long-range dependencies span distant tokens, reflecting the broader sentence context.

Given an input sequence \( H \), we first apply a partial flip (pf) to reverse only the initial sequence segment with dedicated parameter $r$, preserving local patterns. It
selectively reverses the first \( n \) elements while preserving the order of the remaining (\( r = N - n \)) elements, producing a transformed sequence: \([h_n, \dots, h_2, h_1, h_{n+1}, \dots, h_N]\).
This partially flipped input is processed through a Conv1D layer to extract short-range patterns in reverse order:
\begin{equation}
    \tilde{H}^{bwd} = \operatorname{pf}(\operatorname{Conv1d}(\operatorname{pf}(\operatorname{Linear}(\operatorname{DyT}(H)))))
\end{equation}

We flip the output and pass it to an mLSTM to capture long-range reversed dependencies:
\begin{equation}
    {H_t^{bwd}} = \operatorname{ff}(\operatorname{mLSTM}(\operatorname{ff}(\tilde{H}_t^{{bwd}}))) \odot H_t^{norm}
\end{equation}

By processing reversed sequences, the flip-conv-flip-mLSTM pipeline captures both short- and long-range dependencies, preserving aspect-sentiment alignment

\subsection{Multihead Cross Feature Fusion Block (MC2F)}  
To effectively integrate aspect-oriented semantics with long-range and local contextual dependencies in ABSA, we propose MC2F block within AF-MAT framework. Unlike softmax-based attention mechanisms~\cite{tsai-etal-2019}, which are computationally intensive, or Mamba-based models~\cite{HE2025102779}, which depend on CUDA-specific kernels, MC2F is a lightweight, broadly compatible alternative.

MC2F fuses the output of the AA-mLSTM, which emphasizes aspect-relevant semantics (\( H_t^{fwd} \)), with short- and long-range features derived from the reversed FlipMix path (\( H_t^{bwd} \))). This is achieved via an mLSTM-based mechanism, where \( H_t^{fwd} \) serves as queries and keys, and \( H_t^{bwd} \) as values.

The mLSTM fusion incorporates exponentially weighted input and forget gates with stabilization, allowing the model to control memory decay and prioritize relevant past interactions. This enables a soft accumulation of context across timesteps rather than the one-step retrieval of attention. The result is a more temporal-aware and adaptive integration of aspect-oriented, long and local features, aligning closely with the nature of aspect-sentiment interactions that span variable distances in text. The process is defined as: 
\begin{equation}
    {H_t^{com}} = \operatorname{mLSTM}(H_t^{fwd}, H_t^{fwd}, H_t^{bwd}) \odot H_t^{norm}
\end{equation}
The outputs from AA-mLSTM, FlipMix, and MC2F are concatenated to form the final token sequence \( Z \).
\begin{equation}
Z = \operatorname{Linear}\left( H^{fwd} \oplus H^{bwd} \oplus H^{com}\right) + H
\end{equation}

To enable precise sentiment classification in ABSA, we aggregate contextualized embeddings $Z$ using mean pooling to produce a compact representation $H^{mp}$, facilitating downstream tasks. A linear classifier then transforms $H^{mp}$ into logits, which are converted to probabilities via a softmax function, enabling accurate sentiment prediction for aspects. This streamlined pipeline, from embedding to sentiment classification, ensures robust analysis of input text for ABSA tasks.
\begin{equation}
H^{mp} = ( \operatorname{MeanPooling}(Z))
\end{equation}
\begin{equation}
p(a) = \operatorname{Softmax}(W_p H^{mp} + b_p)
\end{equation}

\subsection{Training}
To optimize our AF-MAT model for ABSA, we employ the standard cross-entropy loss as the objective function, calculated across all sentence-aspect pairs in the dataset  $D$. For each pair $(s, a)$, where $s$ is the sentence and $a$ is the aspect, we minimize the negative log-likelihood of the predicted sentiment probability $p(a)$. The loss is defined as: 
\begin{equation}
	L(\theta) = -\sum_{(s,a) \in D} \sum_{c \in C} \log p(a)
\end{equation}
where $\theta$ represents all trainable parameters, and ( C ) denotes the set of sentiment polarity classes. This formulation ensures robust training for precise ABSA.

\section{Experiment}

\subsection{Datasets}
We evaluate our model on three benchmark datasets for ABSA: the Restaurant and Laptop datasets from SemEval 2014 Task 4 \cite{Pontiki2014}, and the Twitter dataset of social media posts~\cite{Dong2014}. Each aspect is annotated with one of three sentiment polarities: positive, neutral, or negative. Table~\ref{tab:dataset_statistics} shows the datasets statistical details.

\begin{table}[h!]
	\centering
	\begin{tabular}{l l c c c}
		\hline
		\textbf{Dataset} & \textbf{Division} & \textbf{Pos} & \textbf{Neg} & \textbf{Neu} \\\hline
		Rest14 & Train & 2164 & 807 & 637 \\
		& Test  & 727  & 196 & 196 \\\hline
		Laptop14 & Train & 976  & 851 & 455 \\
		& Test  & 337  & 128 & 167 \\\hline
		Twitter & Train & 1507 & 1528 & 3016 \\
		& Test  & 172  & 169 & 336 \\\hline
	\end{tabular}
	\caption{Statistics of three benchmark datasets}
	\label{tab:dataset_statistics}
\end{table}

\subsection{Baselines}
To assess the performance of our model, we conduct an extensive comparison with state-of-the-art (SOTA) baselines, encompassing attention-based models, including ATAE-LSTM \cite{Wang2016}, IAN \cite{Ma2017}, RAM \cite{Peng2017}, MGAN \cite{Fan2018}, BERT \cite{Devlin2018}, AEN \cite{Song2019}, GANN \cite{Liu2020}, BiVSNP \cite{ZHU2025125295}, attention-based GRU \cite{Yadav2021}, as well as GCN-based models, such as, CDT \cite{Sun2019}, ASGCN \cite{Zhang2019b}, DGEDT \cite{Tang2020}, DGGCN \cite {Liu2023}, KDGN \cite{Wu2023}, EK-GCN \cite{Gu2023a},  KHGCN \cite{Song2024}, DPWAFGCN-BERT \cite{Yu2025}, IA-GCN \cite{Wu2026}, and MambaForGCN \cite{lawan-etal-2025-enhancing}.

\begin{table*}[t]
\centering
\caption{Experimental results comparison on three publicly available datasets. The best results are highlighted in boldface, and lacking results are marked as “–”.}
\label{tab:experimental_results}
\resizebox{\textwidth}{!}{%
\begin{tabular}{l cc cc cc}
\toprule
\textbf{Model} & \multicolumn{2}{c}{\textbf{Restaurant14}} & \multicolumn{2}{c}{\textbf{Laptop14}} & \multicolumn{2}{c}{\textbf{Twitter}} \\
& \textbf{Acc.} & \textbf{F1} & \textbf{Acc.} & \textbf{F1} & \textbf{Acc.} & \textbf{F1} \\
\midrule
ATAE-LSTM~\cite{Wang2016} & 77.20 & - & 68.70 & - & - & - \\
IAN~\cite{Ma2017} & 78.60 & - & 72.10 & - & - & - \\
RAM~\cite{Peng2017} & 80.23 & 70.80 & 74.49 & 71.35 & 69.36 & 67.30 \\
MGAN~\cite{Fan2018} & 81.25 & 71.94 & 75.39 & 72.47 & 72.54 & 70.81 \\
AEN~\cite{Song2019} & 80.98 & 72.14 & 73.51 & 69.04 & 72.83 & 69.81 \\
GANN~\cite{Liu2020} & 79.70 & - & 72.90 & - & 70.50 & - \\
Attention-based GRU~\cite{Yadav2021} & 81.37 & 72.06 & 75.39 & 70.50 & - & - \\
BiVSNP~\cite{ZHU2025125295} & 82.15 & 71.13 & 75.85 & 71.23 & 73.73 & 70.52 \\
CDT~\cite{Sun2019} & 82.30 & 74.02 & 77.19 & 72.99 & 74.66 & 73.66 \\
ASGCN~\cite{Zhang2019b} & 80.86 & 72.19 &  74.14 & 69.24 & 71.53 & 69.68 \\
DGEDT~\cite{Tang2020} & 83.90 & 75.10 & 76.80 & 72.30 & 74.80 & 73.40 \\
MambaForGCN~\cite{lawan-etal-2025-enhancing} & 84.38 & 77.47 & 78.64 & 76.61 & 75.96 & 74.77 \\
IA-GCN~\cite{Wu2026} & 81.33 & 73.16 & 75.01 & 70.48 & - & - \\
\midrule
\textbf{AF-MAT (ours)} & \textbf{84.67} & \textbf{77.53} & \textbf{78.71} & \textbf{76.66} & \textbf{75.99} & \textbf{74.89} \\
\midrule
BERT~\cite{Devlin2018} & 85.79 & 80.09 & 79.91 & 76.00 & 75.92 & 75.18 \\
KDGN+BERT~\cite{Wu2023} & 87.01 & 81.94 & 81.32 & 77.59 & 77.64 & 75.55 \\
EK-GCN+BERT~\cite{Gu2023a} & 87.65 & \textbf{82.55} & 81.30 & \textbf{79.19} & 75.89 & 75.16 \\
DGGCN+BERT~\cite{Liu2023} & 86.89 & 80.32 & 81.50 & 78.51  & 76.94 & 75.07 \\
KHGCN~\cite{Song2024} & - & - &  80.87 & 77.90 & - & - \\
MambaForGCN+BERT~\cite{lawan-etal-2025-enhancing} & 86.68 & 80.86 & 81.80 & 78.59 & 77.67 & 76.88 \\
DPWAFGCN+BERT~\cite{Yu2025} & 87.32 & 81.47 & 81.66 & 77.51 & - & - \\
\midrule
\textbf{AF-MAT+BERT} & \textbf{87.72} & 81.65 & \textbf{81.87} & 78.74 & \textbf{78.54} & \textbf{76.91} \\
\bottomrule
\end{tabular}
}
\end{table*}

\subsection{Implementation Details}
We initialize word embeddings with 300-dimensional pre-trained GloVe vectors~\cite{Pennington2014}, concatenated with 30-dimensional position and part-of-speech (POS) embeddings. These are input to a BiLSTM model with a hidden size of 50 and a dropout rate of 0.7 to mitigate overfitting. We used 2 layers and 4 heads for AF-MAT+BERT (2 heads for AF-MAT). Model weights are uniformly initialized and optimized using Adam~\cite{Kingma2014} with a learning rate of 0.002 and a batch size of 16 over 50 epochs. For AF-MAT+BERT, BERT derives word representations from its final hidden states. All experiments are implemented in PyTorch and run on an NVIDIA GeForce RTX 4090 GPU with 24 GB of GDDR6X VRAM, CUDA Compute Capability 8.9, driver version 570.144, and CUDA 12.8 support.
\subsection{Main Results}
To demonstrate the effectiveness of AF-MAT, we compare our model with previous works using accuracy and macro-averaged F1 as evaluation metrics, and report results in Table~\ref{tab:experimental_results}. Experimental results show that our AF-MAT model achieves the best performance among non-BERT-based models on the Restaurant14, Laptop14, and Twitter datasets. In particular, our model capture aspect-oriented, local and long-range dependencies between aspect and opinion words, outperforming all other SOTA models. AF-MAT’s superior performance over attention-based models (e.g., ATAE-LSTM, IAN, RAM) and syntax-based models (e.g., CDT, ASGCN) arises from its carefully designed architecture that integrates the AA-mLSTM, FlipMix, and MC2F blocks.  Unlike attention-based models, which are prone to noise in complex or informal sentences like Twitter reviews, or syntax-based models, which are limited by dependency parsing errors, AF-MAT employs gated decay mechanisms that allow selective memory accumulation. This leads to more robust and noise-tolerant sentiment representations. Moreover, AF-MAT enhances aspect-sentiment modeling through stabilized mLSTM gating that filters noise and strengthens the contribution of aspect-relevant information. Its AA-mLSTM introduces an explicit aspect gate, while the FlipMix block captures short- and long-range dependencies from reversed sequences. MC2F further enables efficient fusion of aspect-oriented, local and long features via lightweight multihead gating. These components collectively ensure accurate, efficient, and robust performance, surpassing prior SOTA.

\begin{table*}[h!]
    \centering
    \resizebox{\textwidth}{!}{%
        \begin{tabular}{p{9.5cm} c c c c c}
            \hline
            \textbf{Text} & \textbf{ATAE-LSTM} & \textbf{IAN} & \textbf{IA-GCN} & \textbf{AF-MAT} & \textbf{Labels} \\\hline
            The modern warfare 2 special edition \textcolor{red}{[xbox]}\textsubscript{neg} comes with a 250gb hard drive, holy shit.
            & (O\texttimes) & (O\texttimes) & (O\texttimes & (N\checkmark) & (N) \\\hline

            I opted for the \textcolor{blue}{[SquareTrade 3-year Computer Accidental Protection Warranty]}\textsubscript{pos} which also supports “accidents” like drops and spills that are NOT covered by \textcolor{red}{[AppleCare]}\textsubscript{neg}.
            & (N\texttimes, O\texttimes) & (P\checkmark, P\texttimes) & (P\checkmark, P\texttimes) & (P\checkmark, N\checkmark) & (P, N) \\\hline

            The \textcolor{blue}{[design]}\textsubscript{pos} is very intimate and romantic.
            & (P\checkmark) & (P\checkmark) & (P\checkmark) & (P\checkmark) & (P) \\\hline
            
            Virile heavenly host \textcolor{black}{[nicolas cage]}\textsubscript{neu}.
            & (O\checkmark) & (O\checkmark) & (O\checkmark) & (O\checkmark) & (O) \\\hline
        \end{tabular}%
    }
    \caption{Case studies comparing AF-MAT with SOTA. \checkmark\ denotes correct prediction, and \texttimes\ indicates incorrect prediction. The notations P, N, and O denote positive, negative, and neutral sentiment, respectively.}
    \label{tab:case_study_AF-MAT}
\end{table*}

\subsection{Ablation Study}
To evaluate the effectiveness of AF-MAT’s key components namely: the AA-mLSTM block, the FlipMix block (comprising  pf-Conv1D and ff-mLSTM), and the MC2F fusion block, we conduct a comprehensive ablation study using four model variants: (1) \textit{w/o aspect gate}: removes the aspect-aware gate in AA-mLSTM (replaced by a vanilla mLSTM), (2) \textit{w/o pf-Conv1D}: disables the partial flipping in the pf-Conv1D block, (3) \textit{w/o ff-mLSTM}: omits the full-flip operation in ff-mLSTM and treating pf-Conv1D output as the final backward path, and (4) \textit{w/o MC2F}: removes the MC2F fusion block and instead directly concatenating the outputs of AA-mLSTM and FlipMix.
We evaluate all variants across the three benchmark datasets, with results reported in Table~\ref{tab:ablation_study}. The findings reveal the following insights: (1) The removal of the aspect gate consistently reduces accuracy across datasets, indicating that explicit conditioning on the aspect term plays a vital role in guiding the model’s memory update mechanism, (2) The FlipMix pathway, which sequentially applies pf-Conv1D for short-range dependencies and ff-mLSTM for long-range reasoning, provides complementary temporal views of the sentence. When either the partial or full-flip operation is removed, performance noticeably drops. This confirms that capturing both short- and long-distance dependencies is essential for nuanced sentiment interpretation, (3) The MC2F fusion mechanism proves critical for effective integration of forward and backward features. Unlike simple concatenation, which treats all features equally, MC2F applies a multihead gating strategy through mLSTM to weigh and align information across directions. Its removal significantly degrades performance, highlighting the importance of adaptive feature fusion in AF-MAT.

\subsection{Effect of Hyperparameter r}
In this section, we investigate the impact of \( r \) hyperparameter in the AF-MAT framework using the Restaurant dataset. The results are presented in Figure~\ref{fig:pf_covid1d}.
From the plots, we observe that AF-MAT achieves optimal performance when \( r = 6 \), providing two valuable insights:  
(i) When \( r \) is too large (e.g., \( r = N \), the model struggles to capture directional cues from partial reversal. (ii) When \( r \) is too small (e.g., \( r = 0 \), the model may lose critical short-range aspect-opinion patterns. These findings highlight the importance of selecting a balanced partial flip length in \textit{pf-Conv1D}.

\begin{table}[H]
	\centering
	\renewcommand{\arraystretch}{1.2} % Increase row height slightly for better readability
	\resizebox{\columnwidth}{!}{ % Ensure the table fits perfectly within one column
		\begin{tabular}{@{}lccc@{}} % Remove left and right borders
			\hline
			\textbf{Model} & \textbf{Rest14 Acc.} & \textbf{Lapt14 Acc.} & \textbf{Twit Acc.} \\\hline
			AF-MAT & 84.67 & 78.71 & 75.99 \\
			w/o aspect gate & 83.23 & 77.45 & 74.33 \\
			w/o pf-conv1d & 83.69 & 77.70 & 74.79 \\
                w/o ff-mLSTM & 82.55 & 76.89 & 74.12 \\
			w/o MC2F & 83.11 & 76.97 & 74.24 \\\hline
		\end{tabular}
	}
	\caption{Results of an ablation study (\%)}
	\label{tab:ablation_study}
\end{table}

\subsection{Effect of Hyperparameter L}
\begin{figure}[ht]
    \centering
    \begin{minipage}[t]{0.48\columnwidth}
        \centering
        \includegraphics[width=\linewidth]{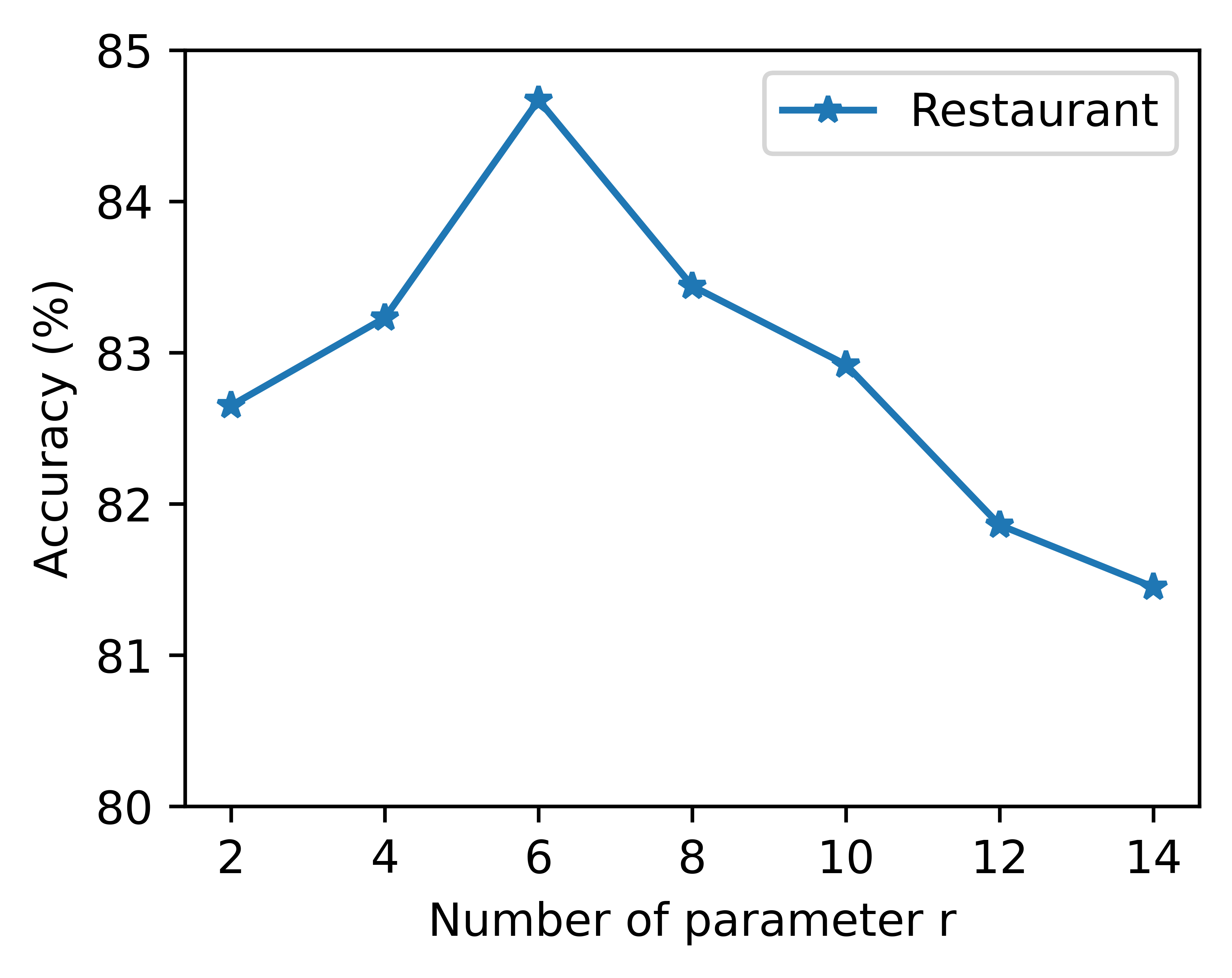}
        \caption{Effect of parameter $r$}
        \label{fig:pf_covid1d}
    \end{minipage}
    \hfill
    \begin{minipage}[t]{0.48\columnwidth}
        \centering
        \includegraphics[width=\linewidth]{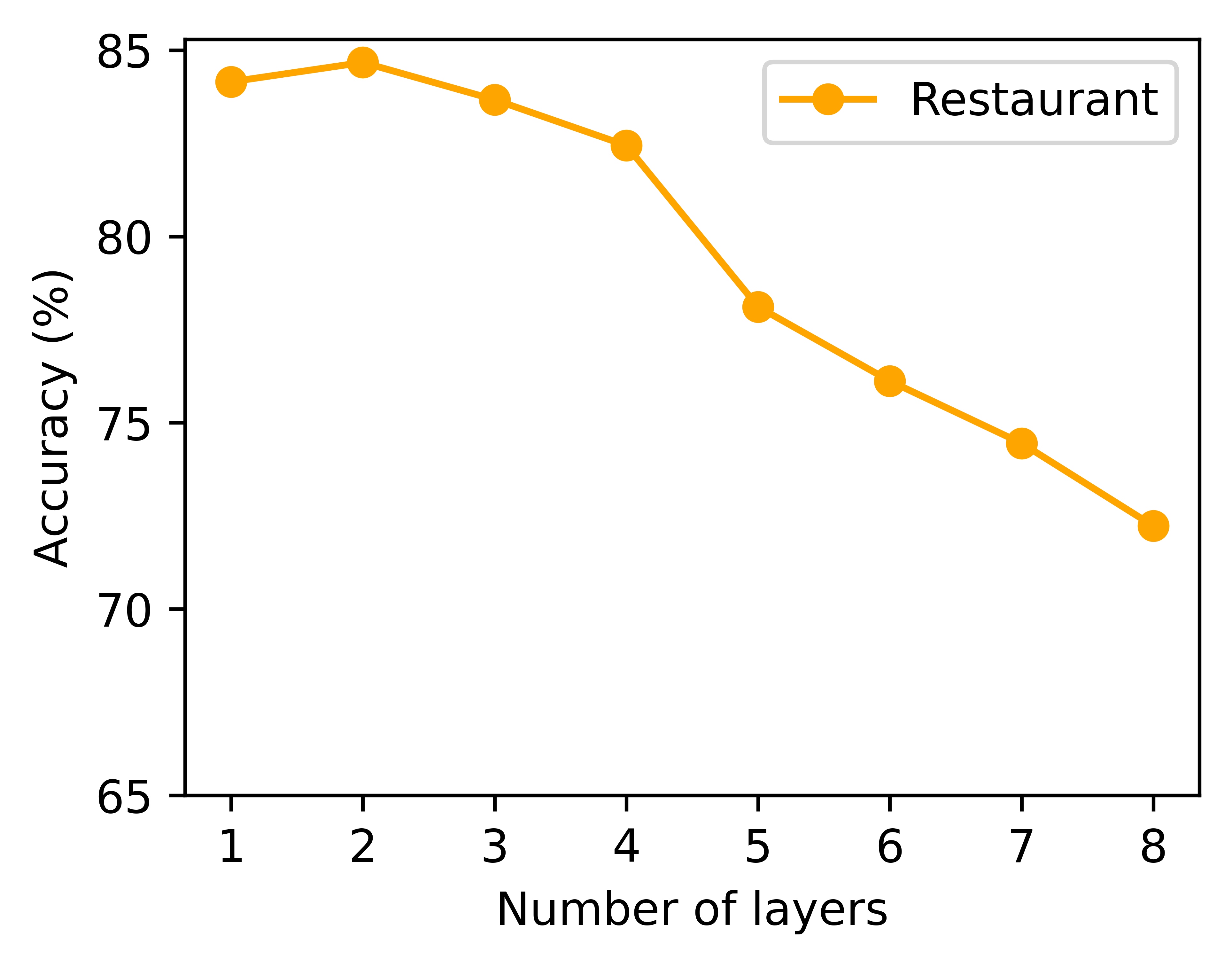}
        \caption{Effect of number of AF-MAT layers}
        \label{fig:ff_mlstmpath}
    \end{minipage}
\end{figure}

In our empirical analysis, illustrated in Figure~\ref{fig:ff_mlstmpath}, we find that the AF-MAT model achieves its highest performance on the Restaurant dataset when configured with two layers. Increasing the number of layers beyond the optimal threshold introduces overfitting and the accumulation of redundant features, ultimately degrading performance. This underscores the importance of careful hyperparameter tuning to identify the ideal layer depth.

\subsection{Case Study}
The first two examples in Table~\ref{tab:case_study_AF-MAT} highlight the importance of integrating both aspect-aware modeling and multi-scale dependency capture in ABSA, particularly in sentences with implicit or contrastive sentiment cues. 

In the first example, \textit{“The modern warfare 2 special edition [xbox] comes with a 250gb hard drive, holy shit”}, all baseline models incorrectly assign O to the aspect \textit{xbox}, likely due to the misleadingly positive tone of the distant ending clause (“holy shit”). This highlights a common challenge in ABSA: an overreliance on local or sentence-level sentiment cues, which often leads to misclassification when the true sentiment is embedded in distant or non-adjacent context. In contrast, AF-MAT effectively models long-range dependencies, correctly identifying the sentiment as negative by linking the aspect \textit{xbox} to the earlier product reference and discounting the emotionally charged but irrelevant clause. This demonstrates AF-MAT’s ability to capture aspect-specific cues across long textual spans, ensuring sentiment is attributed based on the aspect’s true contextual relevance rather than nearby misleading expressions.

The second example, \textit{``I opted for the SquareTrade ... NOT covered by AppleCare''}, the model must infer two contrasting sentiments: positive for \textit{SquareTrade} and negative for \textit{AppleCare}. ATAE-LSTM fails to capture either, while IAN and IA-GCN only partially succeed—predicting \textit{SquareTrade} correctly but mislabeling \textit{AppleCare} as positive. These errors suggest a lack of long-range reasoning and insufficient modeling of negation. AF-MAT, however, correctly predicts both, leveraging its FlipMix block to capture short-range cues (e.g., negation near \textit{AppleCare}) and long-range context (positive framing of \textit{SquareTrade}). This shows the model's strength in parsing subtle contrastive semantics rooted in both local structure and global sentence meaning.

\section{Conclusion}

In this work, we introduced AF-MAT—a novel framework for ABSA that balances accuracy and efficiency through three innovations: an AA-mLSTM that injects inductive bias toward aspect-relevant information via a dedicated gating mechanism, a FlipMix block combining partial and full sequence reversals to jointly capture short- and long-range dependencies, and MC2F, a lightweight multihead mLSTM-based fusion module for adaptive integration of contextual features from both forward and reversed paths. Extensive experiments on three benchmark ABSA datasets confirm that AF-MAT consistently outperforms strong baselines, including attention-based and syntax-aware models, while remaining free from hardware-specific constraints like CUDA.


\begin{thebibliography}{50}
\providecommand{\natexlab}[1]{#1}

\bibitem[{Alkin et~al.(2025)Alkin, Beck, Pöppel, Hochreiter, and Brandstetter}]{alkin2025}
Alkin, B.; Beck, M.; Pöppel, K.; Hochreiter, S.; and Brandstetter, J. 2025.
\newblock Vision-LSTM: xLSTM as Generic Vision Backbone.
\newblock arXiv:2406.04303.

\bibitem[{Beck et~al.(2024)Beck, Pöppel, Spanring, Auer, Prudnikova, Kopp, Klambauer, Brandstetter, and Hochreiter}]{Beck2024}
Beck, M.; Pöppel, K.; Spanring, M.; Auer, A.; Prudnikova, O.; Kopp, M.; Klambauer, G.; Brandstetter, J.; and Hochreiter, S. 2024.
\newblock xLSTM: Extended Long Short-Term Memory.

\bibitem[{Devlin et~al.(2018)Devlin, Chang, Lee, and Toutanova}]{Devlin2018}
Devlin, J.; Chang, M.-W.; Lee, K.; and Toutanova, K. 2018.
\newblock BERT: Pre-training of Deep Bidirectional Transformers for Language Understanding.
\newblock In \emph{Proceedings of NAACL-HLT 2019}.

\bibitem[{Dong et~al.(2014)Dong, Wei, Tan, Tang, Zhou, and Xu}]{Dong2014}
Dong, L.; Wei, F.; Tan, C.; Tang, D.; Zhou, M.; and Xu, K. 2014.
\newblock Adaptive Recursive Neural Network for Target-dependent Twitter Sentiment Classification.
\newblock In \emph{Proceedings of the 52nd Annual Meeting of the Association for Computational Linguistics}, 49--54. Association for Computational Linguistics.

\bibitem[{Fan, Feng, and Zhao(2018)}]{Fan2018}
Fan, F.; Feng, Y.; and Zhao, D. 2018.
\newblock Multi-grained Attention Network for Aspect-Level Sentiment Classification.
\newblock In \emph{Proceedings of the 2018 Conference on Empirical Methods in Natural Language Processing}, 3433--3442.

\bibitem[{Feng et~al.(2022)Feng, Wang, Yang, and Ouyang}]{Feng2022}
Feng, S.; Wang, B.; Yang, Z.; and Ouyang, J. 2022.
\newblock Aspect-based sentiment analysis with attention-assisted graph and variational sentence representation.
\newblock \emph{Knowledge-Based Systems}, 258.

\bibitem[{Gu and Dao(2023)}]{Gu2023}
Gu, A.; and Dao, T. 2023.
\newblock Mamba: Linear-Time Sequence Modeling with Selective State Spaces.

\bibitem[{Gu et~al.(2020)Gu, Dao, Ermon, Rudra, and Ré}]{Gu2020}
Gu, A.; Dao, T.; Ermon, S.; Rudra, A.; and Ré, C. 2020.
\newblock HiPPO: Recurrent Memory with Optimal Polynomial Projections.
\newblock In \emph{34th Conference on Neural Information Processing Systems (NeurIPS 2020)}.

\bibitem[{Gu, Goel, and Ré(2021)}]{Gu2021}
Gu, A.; Goel, K.; and Ré, C. 2021.
\newblock Efficiently Modeling Long Sequences with Structured State Spaces.
\newblock In \emph{ICLR 2022}.

\bibitem[{Gu et~al.(2023)Gu, Zhao, He, Li, and Ying}]{Gu2023a}
Gu, T.; Zhao, H.; He, Z.; Li, M.; and Ying, D. 2023.
\newblock Integrating external knowledge into aspect-based sentiment analysis using graph neural network.
\newblock \emph{Knowledge-Based Systems}, 259.

\bibitem[{Hazarika et~al.(2018)Hazarika, Poria, Vij, Krishnamurthy, Cambria, and Zimmermann}]{hazarika2018}
Hazarika, D.; Poria, S.; Vij, P.; Krishnamurthy, G.; Cambria, E.; and Zimmermann, R. 2018.
\newblock Modeling Inter-Aspect Dependencies for Aspect-Based Sentiment Analysis.
\newblock In Walker, M.; Ji, H.; and Stent, A., eds., \emph{Proceedings of the 2018 Conference of the North {A}merican Chapter of the Association for Computational Linguistics: Human Language Technologies, Volume 2 (Short Papers)}, 266--270. New Orleans, Louisiana: Association for Computational Linguistics.

\bibitem[{He et~al.(2025{\natexlab{a}})He, Cao, Zhang, Yan, Wang, Li, Xie, Hong, and Zhou}]{HE2025102779}
He, X.; Cao, K.; Zhang, J.; Yan, K.; Wang, Y.; Li, R.; Xie, C.; Hong, D.; and Zhou, M. 2025{\natexlab{a}}.
\newblock Pan-Mamba: Effective pan-sharpening with state space model.
\newblock \emph{Information Fusion}, 115: 102779.

\bibitem[{He et~al.(2025{\natexlab{b}})He, Ni, Zhang, Luo, and Wan}]{He2025}
He, X.; Ni, W.; Zhang, Z.; Luo, H.; and Wan, L. 2025{\natexlab{b}}.
\newblock MSGCN-xLSTM: Efficient Wind Power Forecasting Approach Combining Multi-Scale Graph Convolutional Network and Extended LSTM.
\newblock \emph{IEEE Sensors Journal}.

\bibitem[{Hochreiter and Schmidhuber(1997)}]{6795963}
Hochreiter, S.; and Schmidhuber, J. 1997.
\newblock Long Short-Term Memory.
\newblock \emph{Neural Computation}, 9(8): 1735--1780.

\bibitem[{Kingma and Ba(2014)}]{Kingma2014}
Kingma, D.~P.; and Ba, J. 2014.
\newblock Adam: A Method for Stochastic Optimization.
\newblock In \emph{ICLR 2015}.

\bibitem[{Kong et~al.(2025)Kong, Wang, Nie, Zhou, Zohren, Liang, Sun, and Wen}]{Kong2025}
Kong, Y.; Wang, Z.; Nie, Y.; Zhou, T.; Zohren, S.; Liang, Y.; Sun, P.; and Wen, Q. 2025.
\newblock Unlocking the Power of LSTM for Long Term Time Series Forecasting.
\newblock Technical report.

\bibitem[{Kühne et~al.(2025)Kühne, Østergaard, Jensen, and Tan}]{Khne2025}
Kühne, N.~L.; Østergaard, J.; Jensen, J.; and Tan, Z.-H. 2025.
\newblock xLSTM-SENet: xLSTM for Single-Channel Speech Enhancement.

\bibitem[{Lawan et~al.(2025)Lawan, Pu, Yunusa, Umar, and Lawan}]{lawan-etal-2025-enhancing}
Lawan, A.; Pu, J.; Yunusa, H.; Umar, A.; and Lawan, M. 2025.
\newblock Enhancing Long-range Dependency with State Space Model and Kolmogorov-Arnold Networks for Aspect-based Sentiment Analysis.
\newblock In Rambow, O.; Wanner, L.; Apidianaki, M.; Al-Khalifa, H.; Eugenio, B.~D.; and Schockaert, S., eds., \emph{Proceedings of the 31st International Conference on Computational Linguistics}, 2176--2186. Abu Dhabi, UAE: Association for Computational Linguistics.

\bibitem[{Li, Li, and Xiao(2023)}]{Li2023}
Li, P.; Li, P.; and Xiao, X. 2023.
\newblock Aspect-Pair Supervised Contrastive Learning for aspect-based sentiment analysis.
\newblock \emph{Knowledge-Based Systems}, 274.

\bibitem[{Li et~al.(2021)Li, Chen, Feng, Ma, Wang, and Hovy}]{li-etal-2021-dual-graph}
Li, R.; Chen, H.; Feng, F.; Ma, Z.; Wang, X.; and Hovy, E. 2021.
\newblock Dual Graph Convolutional Networks for Aspect-based Sentiment Analysis.
\newblock In Zong, C.; Xia, F.; Li, W.; and Navigli, R., eds., \emph{Proceedings of the 59th Annual Meeting of the Association for Computational Linguistics and the 11th International Joint Conference on Natural Language Processing (Volume 1: Long Papers)}, 6319--6329. Online: Association for Computational Linguistics.

\bibitem[{Liang et~al.(2022)Liang, Su, Gui, Cambria, and Xu}]{Liang2022}
Liang, B.; Su, H.; Gui, L.; Cambria, E.; and Xu, R. 2022.
\newblock Aspect-based sentiment analysis via affective knowledge enhanced graph convolutional networks.
\newblock \emph{Knowledge-Based Systems}, 235.

\bibitem[{Liu et~al.(2023)Liu, Wu, Li, Lu, Li, Wei, Liu, and Feng}]{Liu2023}
Liu, H.; Wu, Y.; Li, Q.; Lu, W.; Li, X.; Wei, J.; Liu, X.; and Feng, J. 2023.
\newblock Enhancing aspect-based sentiment analysis using a dual-gated graph convolutional network via contextual affective knowledge.
\newblock \emph{Neurocomputing}, 553.

\bibitem[{Liu and Shen(2020)}]{Liu2020}
Liu, N.; and Shen, B. 2020.
\newblock Aspect-based sentiment analysis with gated alternate neural network.
\newblock 188: 105010.

\bibitem[{Liu et~al.(2025)Liu, Liu, Wang, Wang, Jia, Wang, Liu, Chang, and Zhao}]{Liu_2025}
Liu, Z.; Liu, Q.; Wang, Y.; Wang, W.; Jia, P.; Wang, M.; Liu, Z.; Chang, Y.; and Zhao, X. 2025.
\newblock SIGMA: Selective Gated Mamba for Sequential Recommendation.
\newblock \emph{Proceedings of the AAAI Conference on Artificial Intelligence}, 39(12): 12264--12272.

\bibitem[{Ma et~al.(2017)Ma, Li, Zhang, and Wang}]{Ma2017}
Ma, D.; Li, S.; Zhang, X.; and Wang, H. 2017.
\newblock Interactive Attention Networks for Aspect-Level Sentiment Classification.
\newblock In \emph{IJCAI'17: Proceedings of the 26th International Joint Conference on Artificial Intelligence}.

\bibitem[{Ouyang et~al.(2024)Ouyang, Xuan, Wang, and Yang}]{Ouyang2024}
Ouyang, J.; Xuan, C.; Wang, B.; and Yang, Z. 2024.
\newblock Aspect-based sentiment classification with aspect-specific hypergraph attention networks.
\newblock \emph{Expert Systems with Applications}, 248: 123412.

\bibitem[{Peng et~al.(2017)Peng, Zhongqian, Lidong, and Yang}]{Peng2017}
Peng, C.; Zhongqian, S.; Lidong, B.; and Yang, W. 2017.
\newblock Recurrent Attention Network on Memory for Aspect Sentiment Analysis.
\newblock In \emph{Proceedings of the 2017 Conference on Empirical Methods in Natural Language Processing}, 452--461.

\bibitem[{Pennington, Socher, and Manning(2014)}]{Pennington2014}
Pennington, J.; Socher, R.; and Manning, C.~D. 2014.
\newblock GloVe: Global Vectors for Word Representation.

\bibitem[{Pontiki et~al.(2014)Pontiki, Papageorgiou, Galanis, Androutsopoulos, Pavlopoulos, and Manandhar}]{Pontiki2014}
Pontiki, M.; Papageorgiou, H.; Galanis, D.; Androutsopoulos, I.; Pavlopoulos, J.; and Manandhar, S. 2014.
\newblock SemEval-2014 Task 4: Aspect Based Sentiment Analysis.
\newblock In \emph{Proceedings of the 8th International Workshop on Semantic Evaluation}, 27--35.

\bibitem[{Song et~al.(2024)Song, Ling, Tu, and Chen}]{Song2024}
Song, X.; Ling, G.; Tu, W.; and Chen, Y. 2024.
\newblock Knowledge-Guided Heterogeneous Graph Convolutional Network for Aspect-Based Sentiment Analysis.
\newblock \emph{Electronics (Switzerland)}, 13.

\bibitem[{Song et~al.(2019)Song, Wang, Jiang, Liu, and Rao}]{Song2019}
Song, Y.; Wang, J.; Jiang, T.; Liu, Z.; and Rao, Y. 2019.
\newblock Attentional Encoder Network for Targeted Sentiment Classification.

\bibitem[{Sun et~al.(2019)Sun, Zhang, Mensah, Mao, and Liu}]{Sun2019}
Sun, K.; Zhang, R.; Mensah, S.; Mao, Y.; and Liu, X. 2019.
\newblock Aspect-Level Sentiment Analysis Via Convolution over Dependency Tree.
\newblock In \emph{Proceedings of the 2019 Conference on Empirical Methods in Natural Language Processing and the 9th International Joint Conference on Natural Language Processing}, 5679--5688.

\bibitem[{Tang, Qin, and Liu(2016)}]{Tang2016}
Tang, D.; Qin, B.; and Liu, T. 2016.
\newblock Aspect Level Sentiment Classification with Deep Memory Network.
\newblock In \emph{Proceedings of the 2016 Conference on Empirical Methods in Natural Language Processing}, 214--224.

\bibitem[{Tang et~al.(2020)Tang, Ji, Li, and Zhou}]{Tang2020}
Tang, H.; Ji, D.; Li, C.; and Zhou, Q. 2020.
\newblock Dependency Graph Enhanced Dual-transformer Structure for Aspect-based Sentiment Classification.
\newblock In \emph{Proceedings of the 58th Annual Meeting of the Association for Computational Linguistic}, 6578--6588.

\bibitem[{Tay, Tuan, and Hui(2018)}]{Tay2018}
Tay, Y.; Tuan, L.~A.; and Hui, S.~C. 2018.
\newblock Learning to attend via word-aspect associative fusion for aspect-based sentiment analysis.
\newblock In \emph{Proceedings of the Thirty-Second AAAI Conference on Artificial Intelligence and Thirtieth Innovative Applications of Artificial Intelligence Conference and Eighth AAAI Symposium on Educational Advances in Artificial Intelligence}, AAAI'18/IAAI'18/EAAI'18. AAAI Press.
\newblock ISBN 978-1-57735-800-8.

\bibitem[{Tsai et~al.(2019)Tsai, Bai, Liang, Kolter, Morency, and Salakhutdinov}]{tsai-etal-2019}
Tsai, Y.-H.~H.; Bai, S.; Liang, P.~P.; Kolter, J.~Z.; Morency, L.-P.; and Salakhutdinov, R. 2019.
\newblock Multimodal Transformer for Unaligned Multimodal Language Sequences.
\newblock In Korhonen, A.; Traum, D.; and M{\`a}rquez, L., eds., \emph{Proceedings of the 57th Annual Meeting of the Association for Computational Linguistics}, 6558--6569. Florence, Italy: Association for Computational Linguistics.

\bibitem[{Wang et~al.(2021)Wang, Tang, Yang, and Wang}]{Wang2021}
Wang, X.; Tang, M.; Yang, T.; and Wang, Z. 2021.
\newblock A novel network with multiple attention mechanisms for aspect-level sentiment analysis.
\newblock \emph{Knowledge-Based Systems}, 227.

\bibitem[{Wang et~al.(2016)Wang, Huang, Zhao, and Zhu}]{Wang2016}
Wang, Y.; Huang, M.; Zhao, L.; and Zhu, X. 2016.
\newblock Attention-based LSTM for Aspect-level Sentiment Classification.
\newblock In \emph{Proceedings of the 2016 Conference on Empirical Methods in Natural Language Processing}, 606--615.

\bibitem[{Wu, Huang, and Deng(2023)}]{Wu2023}
Wu, H.; Huang, C.; and Deng, S. 2023.
\newblock Improving aspect-based sentiment analysis with Knowledge-aware Dependency Graph Network.
\newblock \emph{Information Fusion}, 92: 289--299.

\bibitem[{Wu and Deng(2026)}]{Wu2026}
Wu, Y.; and Deng, G. 2026.
\newblock Aspect-level sentiment analysis based on graph convolutional networks and interactive aggregate attention.
\newblock \emph{Computer Speech and Language}, 95.

\bibitem[{Wu et~al.(2024)Wu, Ma, Lian, Lin, and Zhang}]{Wu2024}
Wu, Z.; Ma, X.; Lian, R.; Lin, Z.; and Zhang, W. 2024.
\newblock CDXFormer: Boosting Remote Sensing Change Detection with Extended Long Short-Term Memory.

\bibitem[{Yadav et~al.(2021)Yadav, Jiao, Goodwin, and Granmo}]{Yadav2021}
Yadav, R.~K.; Jiao, L.; Goodwin, M.; and Granmo, O.~C. 2021.
\newblock Positionless aspect based sentiment analysis using attention mechanism[Formula presented].
\newblock \emph{Knowledge-Based Systems}, 226.

\bibitem[{Yang et~al.(2019)Yang, Zhang, Jiang, and Li}]{Yang2019}
Yang, C.; Zhang, H.; Jiang, B.; and Li, K. 2019.
\newblock Aspect-based sentiment analysis with alternating coattention networks.
\newblock \emph{Information Processing and Management}, 56: 463--478.

\bibitem[{Yu, Cao, and Yang(2025)}]{Yu2025}
Yu, B.; Cao, C.; and Yang, Y. 2025.
\newblock Dynamic position weighting aspect-focused graph convolutional network for aspect-based sentiment analysis.
\newblock \emph{Journal of Supercomputing}, 81.

\bibitem[{Zhang, Li, and Song(2019)}]{Zhang2019b}
Zhang, C.; Li, Q.; and Song, D. 2019.
\newblock Aspect-based Sentiment Classification with Aspect-specific Graph Convolutional Networks.
\newblock In Inui, K.; Jiang, J.; Ng, V.; and Wan, X., eds., \emph{Proceedings of the 2019 Conference on Empirical Methods in Natural Language Processing and the 9th International Joint Conference on Natural Language Processing (EMNLP-IJCNLP)}, 4568--4578. Hong Kong, China: Association for Computational Linguistics.

\bibitem[{Zhang et~al.(2025)Zhang, Zhang, Liu, Xiao, Qian, Ahmed, Ambikairajah, Li, and Epps}]{10985910}
Zhang, X.; Zhang, Q.; Liu, H.; Xiao, T.; Qian, X.; Ahmed, B.; Ambikairajah, E.; Li, H.; and Epps, J. 2025.
\newblock Mamba in Speech: Towards an Alternative to Self-Attention.
\newblock \emph{IEEE Transactions on Audio, Speech and Language Processing}, 33: 1933--1948.

\bibitem[{Zhang, Zhou, and Wang(2022)}]{zhang-etal-2022-ssegcn}
Zhang, Z.; Zhou, Z.; and Wang, Y. 2022.
\newblock {SSEGCN}: Syntactic and Semantic Enhanced Graph Convolutional Network for Aspect-based Sentiment Analysis.
\newblock In Carpuat, M.; de~Marneffe, M.-C.; and Meza~Ruiz, I.~V., eds., \emph{Proceedings of the 2022 Conference of the North American Chapter of the Association for Computational Linguistics: Human Language Technologies}, 4916--4925. Seattle, United States: Association for Computational Linguistics.

\bibitem[{Zhu, Yi, and Luo(2025)}]{ZHU2025125295}
Zhu, C.; Yi, B.; and Luo, L. 2025.
\newblock Aspect-based sentiment analysis via bidirectional variant spiking neural P systems.
\newblock \emph{Expert Systems with Applications}, 259: 125295.

\bibitem[{Zhu et~al.(2025)Zhu, Chen, He, LeCun, and Liu}]{Zhu_2025_CVPR}
Zhu, J.; Chen, X.; He, K.; LeCun, Y.; and Liu, Z. 2025.
\newblock Transformers without Normalization.
\newblock In \emph{Proceedings of the Computer Vision and Pattern Recognition Conference (CVPR)}, 14901--14911.

\bibitem[{Zhu et~al.(2024)Zhu, Liao, Zhang, Wang, Liu, and Wang}]{Zhu2024}
Zhu, L.; Liao, B.; Zhang, Q.; Wang, X.; Liu, W.; and Wang, X. 2024.
\newblock Vision Mamba: Efficient Visual Representation Learning with Bidirectional State Space Model.

\end{thebibliography}
\end{document}